# Optical character recognition quality affects perceived usefulness of historical newspaper clippings

Kimmo Kettunen[1], Heikki Keskustalo[2], Sanna Kumpulainen[2], Tuula Pääkkönen[3] and Juha Rautiainen[3]

[1] *University of Eastern Finland, School of Humanities, Finnish Language and Cultural Research, P.O. Box 111, 80101 Joensuu, Finland*
[2] *Tampere University, Faculty of Information Technology and Communication Sciences, Kalevantie 4, 33014 Tampereen yliopisto, Finland*
[3] *University of Helsinki, The National Library of Finland, Saimaankatu 6, 50100 Mikkeli, Finland*

## Abstract

***Introduction.*** *We study effect of different quality optical character recognition in interactive information retrieval with a collection of one digitized historical Finnish newspaper.*
***Method****. This study is based on the simulated interactive information retrieval work task model. Thirty-two users made searches to an article collection of Finnish newspaper Uusi Suometar 1869–1918 with ca. 1.45 million auto segmented articles. Our article search database had two versions of each article with different quality optical character recognition. Each user performed six pre-formulated and six self-formulated short queries and evaluated subjectively the top-10 results using graded relevance scale of 0–3 without knowing about the optical character recognition quality differences of the otherwise identical articles.*
***Analysis****. Analysis of the user evaluations was performed by comparing mean averages of evaluations scores in user sessions. Differences of query results were detected by analysing lengths of returned articles in pre-formulated and self-formulated queries and number of different documents retrieved overall in these two sessions.*
***Results****. The main result of the study is that improved optical character recognition quality affects perceived usefulness of historical newspaper articles positively.*
***Conclusions****. We were able to show that improvement in optical character recognition quality of documents leads to higher mean relevance evaluation scores of query results in our historical newspaper collection. To the best of our knowledge this simulated interactive user-task is the first one showing empirically that users' subjective relevance assessments are affected by a change in the quality of optically read text.*

Introduction

Digitized versions of historical newspaper collections have been both produced and studied extensively during the last few decades in different parts of the world. In year 2012 it was estimated that digitized newspaper collections in Europe consisted of ca. 129 M pages and 24 000 titles (Dunning, 2012), about 17% of the available content (Gooding, 2018). Since that several national libraries and other stake holders, such as publishers, have produced and are currently producing more and more digitized historical content online out of their newspaper collections. In a recent publication describing in detail ten different digitized historical newspaper collections Beals and Bell (2020) state that

>*over the past thirty years, national libraries, universities and commercial publishers around the world have made available hundreds of millions of pages of historical newspapers through mass digitisation and currently release over one million new pages per month worldwide. These have become vital resources not only for academics but for journalists, politicians, schools, and the general public.*

It is evident that more collections and more data will be available in the future and usage of the collections will increase.

Several projects have studied and developed methods to access the digitized newspaper collections. Out of the projects only a few can be mentioned here. One of the first large European projects was *Europeana* (http://www.europeana-newspapers.eu/, Neudecker & Antonacopoulos, 2016). Europeana gathered a selection of digitized newspaper content produced in different European libraries together, enhanced searchability of the content, developed an evaluation and quality-assessment infrastructure for newspaper digitization, and created best-practice recommendations for newspaper metadata, among other things. Later research projects like *NewsEye* (newseye.eu, Oberblicher et al., 2021), *Impresso* (https://impresso-project.ch/app/), and *Oceanic Exchanges* (https://oceanicexchanges.org/), have studied or are currently studying how to access, develop and enrich the contents of digitized newspaper collections.

In Finland, the most important national project related to the study of digitized historical newspapers has been *Computational History and the Transformation of Public Discourse in Finland, 1640-1910*, which finished at the end of the year 2019. The project used historical newspapers published in Finland during the years 1771–1920 as one of its main research data. One part of the project studied metadata related to the newspapers (Mäkelä et al., 2019; Marjanen et al., 2019), another part of the project produced, e.g., a database out of which reuse of the published news stories could be detected and studied (Salmi et al., 2020). The National Library of Finland participated in the project by providing both data and working with improvement of the data and production of optical character recognition, named entity and article extraction training and evaluation collections (Kettunen, 2019; Koistinen and Kettunen, 2020; Kettunen et al., 2019a).

In this study a 49-year history of Finnish newspaper *Uusi Suometar* is used in a user-oriented study of information retrieval and access. The study can be considered as a realistic approach to information retrieval: participants of the study make searches in an optically read historical newspaper collection to find newspaper articles related to given topics. This is what scholars and lay users of these collections do daily in different parts of the world with varying success. In our evaluation users search the newspaper database both with pre-formulated queries and queries that they formulate on their own based on the topic descriptions. The database of *Uusi Suometar* has two different versions of the content: one with original optical character recognition and one with improved, new optical character recognition. The query engine always searches for the results of queries in the new optical character recognition version of the database and ranks the results according to these. However, retrieved texts presented for reading are balloted in the two different optically read qualities of the same articles. Users of the query system were not aware of differences in the optical character recognition quality when they used the query environment.

The structure of the paper is as follows. The next section describes our research questions and the research method. After that we discuss related research in the chapter *Background.* Newspaper data, its quality and structure are described in the following section, and after that the query interface is described. Sections *Collection and analysis of results* and *Results and their evaluation* describe and analyse the results of the query sessions. Finally, *Discussion,* ends the paper and draws conclusions.

## Research questions and research method

In this interactive information retrieval study, we look answers for the following research questions.

1. Do the search results of pre-formulated queries differ from self-formulated queries of the users?
2. Does different quality of the optical character recognition (old versus new) affect the perceived usefulness of the newspaper clippings? The sub-questions are these two:

   2.1 What happens with perceived usefulness in the case of pre-formulated queries?
   2.2 What happens with perceived usefulness in the case self-formulated queries?

From an information retrieval point of view our research belongs to the tradition of interactive information retrieval, where we use a simulated work task situation approach with its information needs (Borlund, 2000; Ingwersen & Järvelin, 2005, pp. 251–254). Our interactive information retrieval setting uses the three main requirements for this kind of task, as described in Borlund (2000): i) potential users as test persons ii) application of dynamic and individual information needs and iii) use of multidimensional and dynamic relevance judgements. Our interactive information retrieval environment developed for the task is the tool that triggers simulated information needs of the users (cf. section *The query interface*).

Interactive information retrieval approach has the following four main advantages (cf., Borlund, 2000; Borlund and Ingwersen, 1998): first, usage of cover-stories triggers information needs provoked by simulated work tasks. Second, the setting allows individual test persons to assess the usefulness of the newspaper clippings with respect to their own interpretation. Third, using of graded and multidimensional relevance assessments instead of binary and topical ones facilitates both control and repeatability during experimentation by combining the use of both static queries and allowing free-form queries. And finally, the setting enables the use of a realistic search interface with actual data.

To perform the study, we recruited 32 participants for the evaluation task. The student users for the evaluation task were recruited from the courses *Information Retrieval and Language Technology* and *Information Retrieval Methods* at the Tampere University, Faculty of Information Technology and Communication Sciences. Three teachers of information science also participated in the evaluation task. Choice of the participants was based mainly on the ease of getting a large enough group to perform the tasks. A large enough group of real historians would have been harder to gather. We did not collect detailed information about the participants, only the information whether they were students or teachers of information science (cf. Figure 3). Thus, we do not perform any analysis of results based on different user qualities.

The participants, students, and teachers of information science were informed in backgrounding, that they use the information retrieval system of digitized newspaper clippings to write an article of historical events in Finland or world in the time span of 1869-1918. Participants needed to evaluate results of the search in relation to this information need.

Participants were given a one-page instruction leaflet which described the information retrieval task. The leaflet described the general idea of the task and informed the participants about two sessions, gave them the back-grounding simulated work task story, and explained the graded evaluation scale of 0–3. Participants were guided to perform six queries in both sessions with different topics that the query environment balloted one at a time after logging in of the user on. Due to COVID-19 users made searches alone using remote net access to the system, no common session was arranged. Users were instructed to perform the searches on their own schedule in a timeframe of 10 days beginning from March 8, 2021.

The instructions given to the participants as one A4 page gave the participants a background story and formulated the evaluation scale to use in relation to the simulated work task as described in Table 1.

| **The background story** |
|---|
| Imagine that you are writing an article related to topics in history of Finland or world history at the end of 19[th] century or the beginning of 20[th] century. Evaluate quality |

> of the clippings you get as search results. Evaluate the quality of each clipping from the viewpoint, how it helps you to proceed with your article writing.
>
> **Evaluation of the search results (graded relevance scale of 0-3)**
>
> 3. The clipping deals with the topic very broadly and its information content corresponds well with the task. The clipping helps well in accomplishing your task.
> 2. The clipping deals partially with the task or touches it. The content of the clipping helps to some extent in accomplishing your task.
> 1. The clipping does not deal with the actual topic but helps to find better search terms and to limit the topic somehow. It helps indirectly in accomplishing your task.
> 0. The clipping is wholly off topic and does not even help to formulate new queries. This clipping brings no benefit in accomplishing your task.

Table 1: The background story and evaluation instructions given to the participants

The topics were created using history timelines from two popular history encyclopedias: *Suomen historian pikkujättiläinen* ('A small encyclopedia of Finnish history', Zetterberg, 1989) and *Maailmanhistorian pikkujättiläinen ('A small encyclopedia of world history',* Zetterberg, 1988*)*. After finding suitable topics from the timeline, searches to the newspaper data base at digi.kansalliskirjasto.fi were performed to confirm that the database had enough hits related to the topic. During final creation of the query environment many original topics were abandoned, and new ones were created due to too few hits in the final article extraction database. Final topic descriptions were based on Finnish Wikipedia articles related to the topics.

The topics cover the time frame of the historical collection of Uusi Suometar, beginning from 1870s and ending in 1918. First mentioned year in the topic descriptions is 1871, last 1918. Topics cover both domestic and foreign news, the share of domestic news being 21, and foreign 9. Demarcation line between foreign and domestic news is not always sharp, some topics could be classified as both.

Participants performed two separate query sessions: one with pre-formulated queries and another one where they could formulate the queries freely based on the topic descriptions, which were available on the screen of the search environment. In the second session participants could use e.g. Google searches to gather background information about the topics. For each query only the top-10 results were shown to the users. Pre-formulated queries are listed in Appendix 1.

This study uses a relatively small, digitized collection of one newspaper's 49-year history to study users' access to the collection with means of article extraction on the pages of the collection. Article extraction – many times also called segmentation – on historical newspaper collections is not a pervasive property, as good quality separation of articles has been hard to produce automatically (cf. Clausner et al., 2017; Clausner, et. al. 2019; Dengel & Shafait, 2014; Kise, 2014; Jarlbrink & Snickars, 2017). Recent evaluation studies reported in International Conference on Document Analysis and Recognition 2017 and 2019 (Clausner et al., 2017; Clausner et al. 2019) show, that especially state-of-the-art systems (commercial and open source) used many times by libraries in data production do not perform very well in page segmentation and region classification of complex layout document pages. Among the ten digitized historical newspaper collections gathered in Beals and Bell (2020) some do have article extraction, but many do not have it. Article extraction in our study collection has been produced automatically based on a machine learning model and the quality of the result has not been assessed on a large scale, only with a small evaluation collection (cf. section *Newspaper data – Optical Character Recognition quality and article structure*).

Our data has two main challenges for the user, as will be described in more detail later. Shortly put, the optical character recognition quality of the data is not optimal even if it has been improved, and the results of article extraction will make evaluation of query results hard – the texts shown to the users may contain irrelevant content from surrounding context of the newspaper page. These kinds of problems are, however, many times reality with digitized historical newspapers, as testified for example in digital humanities pilot studies of three different digitized newspaper collections in Pfanzelter et al. (2021).

We conduct the study in the framework of interactive information retrieval, but our results are also relevant for the expanding field of digital humanities, which uses increasingly digitized historical newspapers as its research data. We thus try to include some implications for digital humanities in our discussion of the findings of our research.

## Background

Even if new ways of access to the digitized newspaper collections are under development, basic information retrieval tools are still the main entry point to these collections. Users search collections in a query engine interface using keywords to describe their information needs. Information retrieval of historical newspapers has several challenges, among which are e.g., optical character recognition quality, spelling variation of historical language, lack of proper tools for natural language processing of older language, and lack of structure in the optically read documents (Lopresti, 2009; Gotscharek et al., 2011; Piotrowski, 2012; Järvelin et al., 2016; Karlgren et al., 2019; Pfanzelter et al., 2021). In their present state historical newspapers are a hard task to information retrieval engines, and users of the collections, such as researchers of digital humanities, are not very satisfied with the search possibilities and may have low trust on the search results (Jarlbrink and Snickars, 2017; Pfanzelter, et el., 2021).

Textual properties of digitized historical newspapers, such as the quality of optical character recognition and segmentation, are often studied in data-oriented scenarios, which pay attention to the statistical properties of text without consulting the user viewpoint. Effects of sub-quality optical character recognition to efficiency of information retrieval has been studied to fair extent, and the results include both simulations, where quality of the text content has been tampered artificially and original optical character recognition text. Actual user studies in a controlled query-environment, however, have been so far missing. Simulated research settings include e.g., Taghva et al. (1996), Savoy and Naji (2011), and Bazzo et al. (2020), just to mention a few. The general result of these studies is that worse optical character recognition quality lowers query results clearly. Most clear the effect of worse optical character recognition is with short documents and queries of a few words: with these query engine has less evidence for matching of the document and the query words.

Järvelin et al. (2016) report results of information retrieval in a laboratory style collection of digitized Finnish newspapers. Their collection consisted of 180 468 documents (8.45 million pages of newspapers), for which they had developed 56 search topics with graded relevance assessments. Results of the study show, that low level optical character recognition quality of the collection lowered search results clearly, even if heavy fuzzy-matching methods were used in query expansions to improve the results.

If we broaden scope and look at research outside information science, digital humanists have also paid attention to the problems of low-level optical character recognition in digital historical newspaper collections. Jarlbrink and Snickars (2017), for example, show how one digital Swedish newspaper collection, Aftonbladet 1830–1862, *'contains extreme amounts of noise: millions of misinterpreted words generated by OCR, and millions of texts re-edited by the auto-segmentation tool'*. Their main contribution is discussion of low-quality optical character recognition and its effects on using digitized newspapers as research data. Pfanzelter et al. (2021) describe user experiences and needs of digital humanities researchers with three digitized newspaper collections: Austrian ANNO (https://anno.onb.ac.at/), Finnish Digi (digi.kansalliskirjasto.fi), and French Gallica (https://gallica.bnf.fr/) and Retronews (https://gallica.bnf.fr/edit/und/retronews-0). Although their main concern in the paper is related to general functionality demands for interfaces of digitized newspaper collections, they report also experiences related to searchability of the collections. One of their general findings is that *'in some cases, the OCR quality is still very low. After identifying some major issues in this regard, the DH team's reliance on (and trust in) some search results was very low'*. Also, slightly differing opinions have been stated by digital humanities researchers. Strange et al. (2014), for example, state that '*The cleaning was thus desirable but not essential*' referring by cleaning to correction of Optical Character Recognition errors in the digitized texts they were studying.

Van Strien et al. (2020) suggest caution in trusting retrieval results of optically read text. They show, that both rankings of articles and number of returned articles from the query engine are affected by the text quality. Traub et al. (2018) show that better data quality decreases so called retrievability bias, which tends to bring certain documents as search result more often than others (Azzopardi & Vinay, 2008). Chiron et al. (2017) show with respect to the French Gallica collection, that low frequency query words that contain frequent optical character error patterns have a higher risk to result in poor query results.

Traub et al. (2015) interviewed historians to get an impression of their usage of digital archives and their awareness of possible problems related to them. Researchers were usually aware of optical character recognition quality problems of digitized collections and the possible bias caused by quality problems, but they could not quantify, how the problems could affect their research. Traub et al. show also that problems of optical character recognition quality affect different types of research settings differently. They concluded '*that the current knowledge situation on the users' side as well as on the tool makers' and data providers' side is insufficient and needs to be improved*' with respect to the data quality. In the interviews of Korkeamäki and Kumpulainen (2019; Late and Kumpulainen, 2021) historians were asked of their task-based information interaction in digital environments. Some of the interviewed researchers worked a lot with historical optically read texts and some mentioned problems with quality of data especially in the analysis phase of the data: '*Especially when using big data, historians had to evaluate the impact of the noisy OCR on the results and if the analysis is worth doing*' (Korkeamäki and Kumpulainen (2019). Some of interviewed did not trust that search in the optically read collections would retrieve all the relevant hits as results (Late and Kumpulainen, 2021).

### Newspaper data – Optical Character Recognition quality and article structure

Our search collection consists of the whole history of *Uusi Suometar* 1869–1918, ca. 86 000 pages and 306.8 million words (Kettunen and Koistinen, 2020). Uusi Suometar was at the time of its publication one of the most important Finnish language newspapers in Finland, where newspapers were published in two languages, Finnish and Swedish (Hynynen, 2019). The newspaper data has been scanned and optically read as part of the digitization of Finnish newspapers and journals at the National Library of Finland, and is part of the digitized content offered by the National Library of Finland (https://digi.kansalliskirjasto.fi/etusivu?set_language=en)

The original optical character recognition for *Uusi Suometar* was performed using a line of ABBYY FineReader® products. Quality of the digitization of the whole collection of Finnish newspapers 1771–1910 has been estimated in Kettunen and Pääkkönen (2016). They conclude that ca. 70–75% of the words in the Finnish language 2.4-billion-word index database could be recognized by using automatic morphological analysers. In general, this level of word level quality can be considered quite typical for older digitisations of historical newspaper collections (cf. e.g., Tanner et al., 2009).

Improved optical character recognition for the whole history of *Uusi Suometar* was achieved with Tesseract v.3.0.4.01. Improvement to the earlier quality in recognition of words is approximately 15% units as a mean over the whole period. On average 83% of the words of the newspaper were recognized with automatic morphological analysers, and the recognition rate varied from ca. 78 to 88% over the 49 years. For the old Optical Character Recognition mean word recognition rate was 68.2% (Kettunen and Koistinen, 2019; Koistinen et al. 2020). Even if the improvement in Optical Character Recognition quality is considerable, the improved quality can still be challenging for information retrieval engines, especially with short queries and articles, where the information retrieval engine has less evidence for matching the query words and collection data in the engine's index (Järvelin et al., 2016; Mittendorf & Schäuble, 2000). In a recent study Bazzo et al. (2020), for example, found that statistically significant degradation of search results begins already at the word error rate of 5%, when Portuguese pdf texts with artificially induced errors were sought for with state-of-the art modern query engine and algorithms. Same kind of results have been achieved in most of the current studies, as a current survey paper of Nguyen et al. (2021) shows. It should also be emphasized that even if substantially better-quality Optical Character Recognition for collections can be achieved with latest tools like Transkribus (Muehlberger et al., 2019), new Optical Character Recognition of large old historical collections is tedious and time consuming – thus the problem of low-level optically read text will stay for some time.

Newspaper data at the National Library of Finland was originally scanned and recognized page by page without article structure information besides basic layout of the pages. For this study we used articles that were extracted automatically from the pages of *Uusi Suometar* using a trained machine learning model with software PIVAJ (Hebert et al., 2014a, b; Kettunen et al. 2019). The training of the PIVAJ model was based on 168 pages of manually marked data that had different number of columns (varying from 3 to 9). Kettunen et al. (2019) reported success per centages of 67.9, 76.1, and 92.2 for an evaluation data set of 56 pages in three different evaluation scenarios based on Clausner et al. (2011) using layout evaluation software from PRImA (https://www.primaresearch.org/). In the automatic segmentation process the collection of *Uusi Suometar* was divided into 1 459 068 articles with PIVAJ. Figure 1 shows an example of PIVAJ's graphical output for article structure. Different articles of the page are shown with different colours.

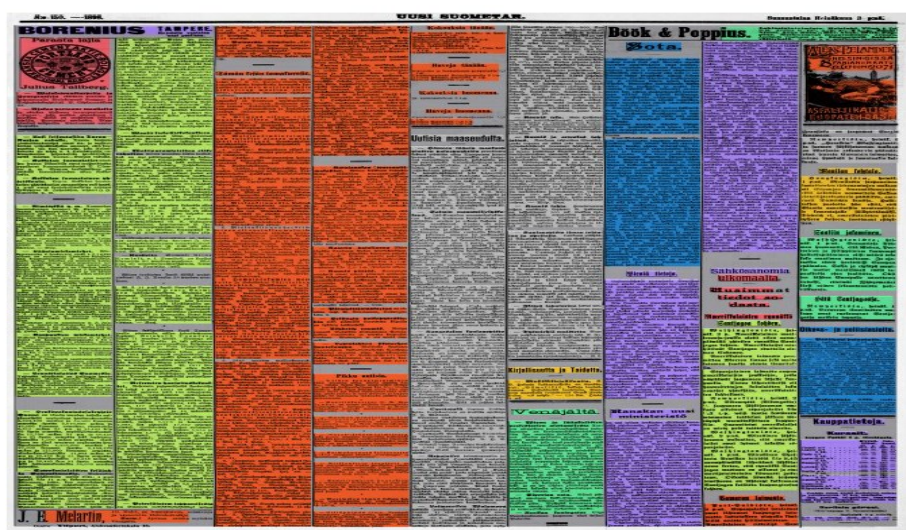

Figure 1: Automatically produced article structure of one page from Uusi Suometar from the graphical output of PIVAJ (Kettunen et al., 2019)

In the article extraction of the whole history of Uusi Suometar article separation is far from optimal, and articles are perhaps best called automatically extracted clippings with varying length. In the search evaluation task, these clippings are documents that users search and evaluate. It should be emphasized, that the article segmentation that was producible for the whole history of Uusi Suometar is experimental and its quality will bring one layer of difficulty to the evaluation of search results. As Jarlbrink and Snickars (2017) formulate it, auto segmentation tools create random texts, and borders of text snippets are fuzzy. This feature was informed to the users in the instructions of the search task.

### The query interface

Participants of the evaluation task performed their task using the query engine *Elastic search* (https://www.elastic.co/), version 7.3.2, which is the background engine of the library's presentation system. Queries were performed in AND mode, where every query term is sought for in the documents. Hits of the search engine shown for the users needed to be at least 500 characters long to avoid very short text passages which would be hard to evaluate. The index of the newspaper collection's database is lemmatized, i.e., it contains base forms of the words, which is crucial as Finnish is a highly inflected language (Järvelin et al., 2016). Search environment and database out of the Uusi

Suometar article extraction results were developed by Evident Solutions Ltd. (https://evident.fi/).

Figure 2 shows the search environment we created for the evaluation of the search results.

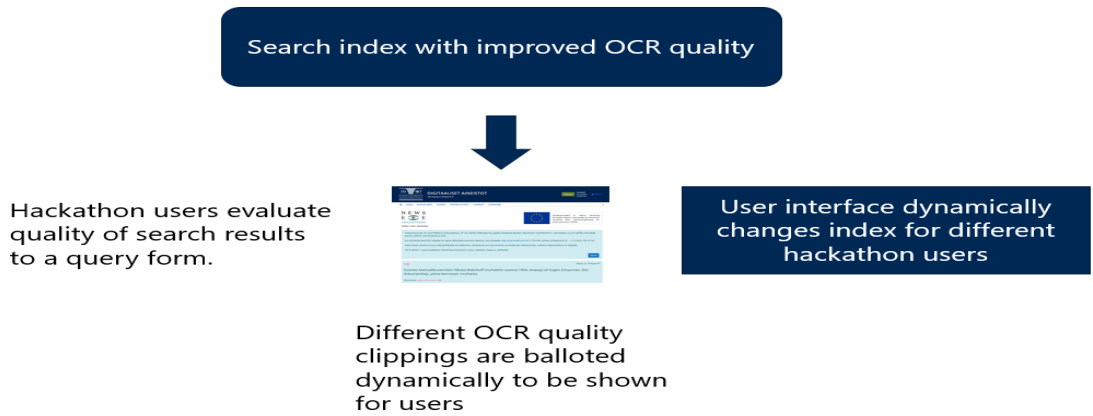

Figure 2: Index side view: improved Optical Character Recognition version of text is always used in the retrieval phase

Figure 3 depicts the query interface the participants used.

Figure 3: Screenshot of the query interface

Figure 3 shows the query interface after a pre-formulated query has been performed and 10 results retrieved. Text on the blue background on the top describes the topic and shows the pre-formulated query beneath in pink. The light purple rectangle below shows the beginning of the first query result. Grading buttons are on the right side of the rectangle. Underneath the text snippet of the result on the left is the button for opening the clipping in its whole. The button also shows the character length of the clipping. Matches of the query words are highlighted in the snippet view and in the actual clipping view which the participants used for evaluating relevance of the clippings.

Collection and analysis of the results

Query results of the user sessions were collected for analysis to a query log, which is shown in Figure 4.

Figure 4: Screenshot of the query log

The columns in the query log indicate the following data beginning from the left: 1) query words 2) session: pre-formulated or self-formulated queries, A and B, respectively 3) number of the topic 4 optical character recognition quality in the results (0 for the old and 1 for the new) 5) user id 6) role of the user (student or teacher) 7) id number of the result clipping 8) user-given evaluation result on the scale of 0-3 9) date and time of the performance 10) possible change for the evaluation: time stamp 11) size of the clipping in characters 12) rank (1-10) of the result clipping in the result list.

The interactive information retrieval system balloted the topics for each user so that out of the 32 users' work we got 3893 evaluations. There were 1861 evaluations of pre-formulated queries and 2032 evaluations of self-formulated queries. Differences in these numbers are due to the fact that part of the users did not perform all their tasks, and some did more than required: if 32 users had each performed 120 evaluations (2 sessions and 10 evaluations for six queries), the log should contain 3840 evaluations evenly divided to the two different sessions.

Analysis of the self-formulated queries showed that users formulated slightly longer queries than the pre-formulated queries were. The mean length of the pre-formulated queries is 2.87 words, and the mean length of the self-formulated queries is 3.15 words, which is a 9.8% increase. The clippings the users evaluated were of varying length. We had set a minimum length of 500 characters for the results to be shown for users, but no maximum length. The mean length of the clippings in all the results of the two sessions was 6116 characters.

Some type of query structuring was used very little in users' self-formulated queries: 111 queries used AND connector and 90 queries connector OR. Combined AND and OR was used in 20 queries. Truncation operators ? and * were both used 43 times in 33 queries.

We assumed that the user interface would take care of the number of queries and evaluations each user finished. However, some of the users did not finish all the queries or evaluations in the pre-formulated queries' session, because the possibility of a user's premature quitting was not taken care of in the system. In the self-formulated queries' session, the users could edit their queries and resubmit them after they had performed them once, and the query log stored all the results for the same query with possible query variations. These issues were unexpected and should be taken care of in the user interface and instructions.

## Results and their evaluation

As the example of the log shows, the result set is multifaceted and could be analysed in many ways. In this study we concentrate on the analysis of the main results of the query sessions to be able to answer our research questions stated in the second section of the study.

Figure 5 shows mean averages for evaluations of the individual queries in both sessions with different quality optical character recognition.

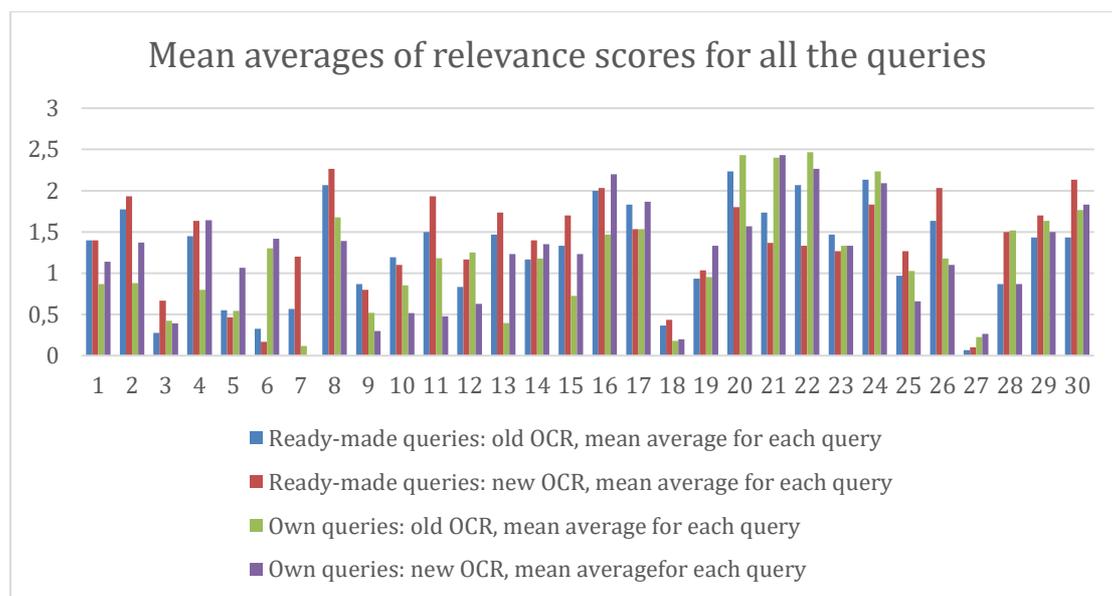

Figure 5: Mean averages of relevance scores for the top-10 clippings retrieved for all topics: graded relevance scale of 0-3 was used

Mean averages for the relevance scores over the whole query set are shown in Table 3.

| Pre-formulated queries, old optical character recognition: mean average for evaluations of the query set | Pre-formulated queries, new optical character recognition: mean average for evaluations of the query set | Self-formulated queries, old optical character recognition: mean average for evaluations of the query set | Self-formulated queries, new optical character recognition: mean average for evaluations of the query set |
|---|---|---|---|
| 1.26 | 1.36 | 1.17 | 1.19 |

Table 2: Mean averages for the relevance scores over the whole query set: graded relevance scale of 0-3 was used

From Figure 5 and Table 2 we can see, that especially pre-formulated queries benefited from improved optical character recognition. The mean average evaluation score for the improved optical character recognition in the pre-formulated queries is 7.93% higher than with the old optical character recognition. In self-formulated queries the difference is clearly smaller: 1.71%. Three queries, #3, #18 and #27, got low evaluations in all the scenarios. Two queries, #6, #7, got low evaluations in part of the scenarios: #6 with pre-formulated queries with both optical character recognition qualities and #7 with self-formulated queries with both optical character recognition qualities.

Inspection of query-per-query results shows that pre-formulated queries get better mean evaluation scores in 19 cases out of 30. There is one tie and 10 queries, where evaluations of old optical character recognition get better mean evaluation scores. With self-formulated queries the result is more even: new optical character recognition results get better mean evaluation scores in 15 cases, and old optical character recognition results get better mean evaluation scores in 14 cases; there is one tie. This is depicted in detail figure 6.

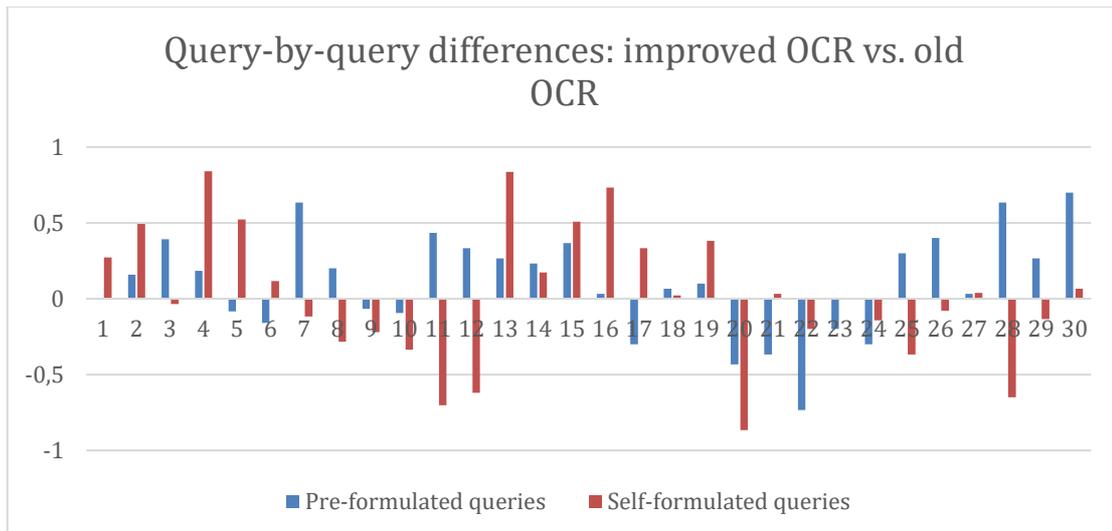

Figure 6: Quey-by-query differences of relevance scores' mean averages for the top-10 clippings: graded relevance scale of 0-3 was used

The mean length of the clippings retrieved by the search engine in the two sessions differs quite a lot. With pre-formulated queries the mean length of the clippings was 5467 characters, and with self-formulated queries 6711 characters, which is a 22.75% difference. It is thus possible that the longer result clippings of the self-formulated queries were harder to evaluate for users and this is reflected in the lower evaluation results, as was seen in Figure 5 and Table 2. With longer clippings the users may get tired or frustrated, which may lower their evaluations. The longer clippings may also be fuzzier than the shorter ones and contain text from adjacent text segments. This aspect would need further study.

In the case of the pre-formulated queries, the difference in the effect of Optical Character Recognition quality on the relevance judgements was statistically significant (p=0.002, Wilcoxon's signed rank test, Croft et al., 2010) when the relevance of the individual underlying documents was judged based on two possible levels of Optical Character Recognition quality. The difference in the overall effectiveness of retrieval (measured via mean average of cumulated gain among top-10 documents in the case of 30 topics), however, was not statistically significant (p=0.10, Wilcoxon's signed rank test).

In the case of self-formulated queries' session, statistically significant differences were not observed. Self-formulated queries retrieved almost four times larger set of unique documents for all user searchers than pre-formulate queries; these queries also differed to some extent from the pre-defined queries (they were slightly longer), and they returned longer documents. However, we did not study these differences and effects deeper.

Discussion

To the best of our knowledge this is the first study showing empirically, based on simulated work task situations (Borlund, 2000), that the subjective relevance assessments of the test persons were affected by the change of quality of the optically read text presented to them. Earlier studies on the effects of Optical Character Recognition quality have been performed in data-oriented settings, often using laboratory-style tests and artificially tampered data or described subjective experiences of users regarding the effects of Optical Character Recognition quality on their work.

The answer to our first research question was, that pre-formulated and self-formulated queries differ in their results. Self-formulated queries of users were slightly longer than the pre-formulated queries that had been created for the topics. Result documents of queries in the two different query sessions differed clearly in length and number. It is possible that the longer result documents of the self-formulated queries were harder to evaluate for users, and they could also contain more text from the adjacent text segments, which would make their evaluation more difficult. This aspect would need more detailed comparison between result clippings of the two sessions.

The answer to our second research question and its first sub-question is clearly positive. With pre-formulated queries the improved optical character recognition quality clippings got a 7.93% better mean relevance score, and the difference to the basic level optical character recognition quality evaluations was statistically significant. Answer to the second sub-question was not as clear. With self-formulated queries we found a small difference of 1.71% in the evaluations in favour of improved optical character recognition. However, this difference was not statistically significant.

Our query environment implementation for the evaluation of two optically read text qualities is a first version of the system. As such it works well, but experience from user sessions showed that it has features that could be developed.

We assumed that the user interface would take care of the number of queries and evaluations each user finished. However, some of the users did not finish all the queries or evaluations in the pre-formulated queries' session, because the possibility of a user's premature quitting was not taken care of in the system. In the self-formulated queries' session, the users could edit their queries after they had performed them, and the query log stored all the results for the same query with possible query variations. These user behaviours were unexpected and should be taken care of in developing the user interface and instructions.

We had also differences in the two query types. In the pre-formulated queries' session, the search was run in Elastic's AND mode, where all the query words were sought for. In the self-formulated queries' session users could use query operators AND, OR and NOT, if they wished. Users' self-formulated queries were also slightly longer than pre-formulated queries. The effect of these differences to the evaluation results is hard to establish conclusively. Evidently it would be more consistent to instruct users not to use any operators in the self-formulated queries' session and instead run the search in AND mode.

The clippings the users evaluated were of varying length. The mean length of the clippings in all the results of the two sessions was 6116 characters. With pre-formulated queries the mean length of the clippings was 5467 characters, and with self-formulated queries 6711 characters – a 22.75% difference. It is thus possible that the longer clipping results of the self-formulated queries were harder to evaluate for users and this is reflected in the lower evaluation results. With longer clippings the users may get tired or frustrated, which may lower their evaluations. Longer clippings may also be fuzzier than shorter ones in their content.

One possible development issue for the evaluation could be evaluation of the clippings' overall textual and segmentation quality by the users. Our article segmentation for the collection is experimental, and many of the clippings may be quite hard to read due to fuzzy boundaries: the clippings may contain text from adjacent segments, which affects evaluations. Combination of relevance evaluation scores and users' scores for the quality of the clipping boundaries might bring new insights to the evaluation and reveal usability of the optically read texts better than usage of traditional relevance assessments only.

The well-known simulated work tasks used in interactive information retrieval have been used in this study to answer the question of optical character recognition quality's effect on relevance evaluation of retrieval results in a Finnish historical newspaper collection. We have shown that improvement in optical character recognition quality of documents leads to higher mean relevance evaluation scores in a simulated work task scenario. This means that perceived usefulness of historical newspaper clippings increases with better optical character recognition quality. Although our results were achieved with one language and one specific collection, our method is generalizable to any language and can be evaluated with further users and different collections.

Our results should be seen both in the context of information retrieval and requirements of digital humanities scholars and lay users of the collections. These results bring more weight to both higher quality document need of digital humanists and efforts of improving quality of optical character recognition with new developments in software. Better quality of optically read historical documents should be strived for both for the sake of research and lay users.

# References

Azzopardi, L., & Vinay, V. (2008). Retrievability: an evaluation measure for higher order information access tasks. In Proceedings of the 17th ACM conference on Information and knowledge management (CIKM '08). Association for Computing Machinery, New York, NY, USA, 561–570. DOI: https://doi.org/10.1145/1458082.1458157

Bazzo G.T., Lorentz G.A., Suarez Vargas D., & Moreira V.P. (2020) Assessing the Impact of OCR Errors in Information Retrieval. In: Jose J. et al. (eds) Advances in Information Retrieval. ECIR 2020. Lecture Notes in Computer Science, vol 12036. Springer, Cham. https://doi.org/10.1007/978-3-030-45442-5_13

Beals, M. H. & Emily Bell, with contributions by Ryan Cordell, Paul Fyfe, Isabel Galina Russell, Tessa Hauswedell, Clemens Neudecker, Julianne Nyhan, Sebastian Padó, Miriam Peña Pimentel, Mila Oiva, Lara Rose, Hannu Salmi, Melissa Terras, and Lorella Viola (2020). The Atlas of Digitised Newspapers and Metadata: Reports from Oceanic Exchanges. Loughborough: 2020. DOI: 10.6084/m9.figshare.11560059.

Borlund, P. (2000). Experimental Components for the Evaluation of Interactive Information Retrieval Systems. Journal of Documentation, 56(1), 71–90.

Borlund, P., & Ingwersen, P. (1998). Measures of relative relevance and ranked half-life: performance indicators for interactive IR. SIGIR '98: Proceedings of the 21st annual international ACM SIGIR conference on Research and development in information retrieval August 1998, 324–331. https://doi.org/10.1145/290941.291019

Chiron, G., Doucet, A., Coustaty, M. Visani, M. & Moreux, J. (2017). Impact of OCR Errors on the Use of Digital Libraries: Towards a Better Access to Information. 2017 ACM/IEEE Joint Conference on Digital Libraries (JCDL), Toronto, ON, 2017, pp. 1-4, doi: 10.1109/JCDL.2017.7991582.

Clausner, C., Pletshacher, S., & Antonacopoulos, A. (2011). Scenario Driven In-Depth Performance Evaluation of Document Layout Analysis Methods. 2011 International Conference on Document Analysis and Recognition (ICDAR). DOI: 10.1109/ICDAR.2011.282

Clausner, C., Pletshacher, S., & Antonacopoulos, A. (2017). ICDAR2017 Competition on Recognition of Documents with Complex Layouts – RDCL2017. https://ieeexplore.ieee.org/document/8270160

Clausner, C., Antonacopoulos, A., & Pletschacher, S. (2019). ICDAR2019 Competition on Recognition of Documents with Complex Layouts – RDCL2019. Proceedings of the 15th International Conference on Document Analysis and Recognition (ICDAR2019), Sydney, Australia, September 2019, pp. 1521-1526

Croft, W. B., Metzler, & D., Strohman, T. (2010). Search Engines. Information Retrieval in Practice. Pearson.

Dengel, A., & Shafait, F. (2014). Analysis of the Logical Layout of Documents. In Doerman, D., Tombre, K. (eds.) Handbook of Document Image Processing and Recognition, 177–222. Springer. DOI 10.1007/978-0-85729-859-1

Dunning, A. (2012). European Newspaper Survey Report. http://www.europeana-newspapers.eu/wp-content/uploads/2012/04/D4.1-Europeana-newspapers-survey-report.pdf

Gooding, P. (2018). Historic Newspapers in the Digital Age. Search All About It! Routledge.

Gotscharek, A., Reffle, U., Ringlstetter, C., Schulz, K., & Neumann, A. (2011). Towards information retrieval on historical document collections: the role of matching procedures and special lexica. IJDAR 14, 159–171. https://doi.org/10.1007/s10032-010-0132-6

Hebert, D., Palfray, T., Nicolas, T., Tranouez, P., & Paquet, T. (2014a). PIVAJ: displaying and augmenting digitized newspapers on the web experimental feedback from the "Journal de Rouen" collection. In: Proceeding DATeCH 2014 Proceedings of the First International Conference on Digital Access to Textual Cultural Heritage, pp. 173–178. ttp://dl.acm.org/citation.cfm?id=2595217

Hebert, D., Palfray, T., Nicolas, T., Tranouez, P., & Paquet, T. (2014b). Automatic article extraction in old newspapers digitized collections. In: Proceeding DATeCH 2014 Proceedings of the First International Conference on Digital Access to Textual Cultural Heritage, pp. 3–8. http://dl.acm.org/citation.cfm?id=2595195

Hynynen, M.-L. (2019.) Building a Bilingual Nation. https://www.newseye.eu/blog/news/building-a-bilingual-nation/

Jansen, B. J., Spink, A., & Saracevic, T. (2000). Real life, real users, and real needs: A study and analysis of user queries on the Web. Information Processing and Management, 36(2), 207–227. https://doi.org/10.1016/S0306-4573(99)00056-4

Jarlbrink, J. & Snickars, P. (2017). Cultural heritage as digital noise: nineteenth century newspapers in the digital archive. Journal of Documentation, Vol. 73 No. 6, pp. 1228-1243. https://doi.org/10.1108/JD-09-2016-0106

Järvelin, A., Keskustalo, H., Sormunen, E., Saastamoinen, M., & Kettunen, K., 2016. Information retrieval from historical newspaper collections in highly inflectional languages: A query expansion approach. *Journal of the Association for Information Science and Technology*, 67(12), 2928-2946.

Järvelin, K., & Ingwersen, P. (2005). The Turn. Integration of Information Seeking and Retrieval in Context. Springer.

Karlgren, J., Hedlund, T., Järvelin, K., Keskustalo, H., & Kettunen, K. (2019). The challenges of language variation in information access. In Nicola Ferro, Carol Peters (eds.), From Multilingual to Multimodal: The Evolution of CLEF over Two Decades. Lessons Learned from 20 Years of CLEF, Springer, 201–216.

Kantor, P.B., & Voorhees, E.M. (2000). The TREC-5 confusion track: comparing retrieval methods for scanned text. Inf. Retrieval 2(2), 165–176.

Kettunen, K., & Koistinen, M. (2019). Open Source Tesseract in Re-OCR of Finnish Fraktur from 19th and Early 20th Century Newspapers and Journals – Collected Notes on Quality Improvement. DHN2019.

Kettunen, K., & Pääkkönen, T. (2016). Measuring lexical quality of a historical Finnish newspaper collection – analysis of garbled OCR data with basic language technology tools and means. LREC 2016, Tenth International Conference on Language Resources and Evaluation. http://www.lrec-conf.org/proceedings/lrec2016/pdf/17_Paper.pdf

Kettunen, K., Ruokolainen, T., Liukkonen, E., Tranouez, P., Anthelme, D., & Paquet, T. (2019a). Detecting Articles in a Digitized Finnish Historical Newspaper Collection 1771–1929: Early Results Using the PIVAJ Software. DATeCH 2019.

Kettunen, K., Pääkkönen, T., & Liukkonen, E. (2019b). Clipping the Page – Automatic Article Detection and Marking Software in Production of Newspaper Clippings of a Digitized Historical Journalistic Collection. In A. Doucet et al. (Eds.), TPDL 2019, LNCS 11799, 356–360.

Kise, K. (2014). Page Segmentation Techniques in Document Analysis. In Doerman, D., Tombre, K. (eds.) Handbook of Document Image Processing and Recognition, 135–175. Springer. DOI 10.1007/978-0-85729-859-1

Koistinen M., Kettunen K., & Kervinen J. (2020). How to Improve Optical Character Recognition of Historical Finnish Newspapers Using Open Source Tesseract OCR Engine – Final Notes on Development and Evaluation. In: Vetulani Z., Paroubek P., Kubis M. (eds) Human Language Technology. Challenges for Computer Science and Linguistics. LTC 2017. Lecture Notes in Computer Science, vol 12598. Springer, Cham. https://doi.org/10.1007/978-3-030-66527-2_2

Korkeamäki, L., & Kumpulainen, S. (2019). Interacting with Digital Documents: A Real Life Study of Historians' Task Processes, Actions and Goals. In Proceedings of the 2019 Conference on Human Information Interaction and Retrieval (CHIIR '19). Association for Computing Machinery, New York, NY, USA, 35–43. DOI:https://doi.org/10.1145/3295750.3298931

Late, E. and Kumpulainen, S. (2021). Interacting with digitised historical newspapers: understanding the use of digital surrogates as primary sources. Journal of Documentation. https://doi.org/10.1108/JD-04-2021-0078.

Lopresti, D. (2009). Optical Character Recognition Errors and Their Effects on Natural Language Processing. International Journal on Document Analysis and Recognition, 12: 141–151. DOI: https://doi.org/10.1007/s10032-009-0094-8

Maailmanhistorian pikkujättiläinen (1988). Seppo Zetterberg et al. (eds). WSOY.

Mäkelä, E., Tolonen, M., Marjanen, J., Kanner, A., Vaara, V., & Lahti, L. (2019). Interdisciplinary collaboration in studying newspaper materiality. In: Krauwer, S. and Fišer, D. (eds.). Proceedings of the Twin Talks Workshop, co-located with Digital Humanities in the Nordic Countries (DHN 2019). Aachen: CEUR Workshop Proceedings vol. 2365: 55–66. http://ceur-ws.org/Vol-2365/07-TwinTalks-DHN2019_paper_7.pdf

Marjanen, J., Vaara, V., Kanner, A., Roivainen, H., Mäkelä, E., Lahti, L., & Tolonen, M. (2019). A National Public Sphere? Analyzing the Language, Location, and Form of Newspapers in Finland, 1771–1917. Journal of European Periodical Studies, 4(1): 54–77. DOI: 10.21825/jeps.v4i1.10483

Mittendorf, E., & Schäuble, P. (2000). Information Retrieval Can Cope with Many Errors. Information Retrieval, 3(3): 189–216. DOI: https://doi.org/10.1023/A:1026564708926

Muehlberger, G., Seaward, L., Terras, M., Oliveira, S. A., Bosch, V., Bryan, M., & al. (2019) Transforming scholarship in the archives through handwritten text recognition: Transkribus as a case study. Journal of Documentation, Vol. 75, No. 5, 2019, pp. 954-976.

Neudecker, C., & Antonacopoulos, A. (2016). Making Europe's Historical Newspapers Searchable. 12th IAPR Workshop on Document Analysis Systems (DAS), Santorini, Greece, 2016, pp. 405-410, doi: 10.1109/DAS.2016.83.

Nguyen, T., Jatowt, A., Coustaty, M. & Doucet, A. (2021). Survey of Post-OCR Processing Approaches. ACM Comput. Surv. 54, 6, Article 124 (July 2021), 37 pages, DOI: https://doi.org/10.1145/3453476.

Oberbichler, S., Boroş, E., Doucet, A., Marjanen, J., Pfanzelter, E., Rautiainen, J., Toivonen, H., & Tolonen, M. (2021). Integrated interdisciplinary workflows for research on historical newspapers: Perspectives from humanities scholars, computer scientists, and librarians. Journal of the Association for Information Science and Technology, 1–15. https://doi.org/10.1002/asi.24565

Pfanzelter, E., Oberbichler, S., Marjanen, J., Langlais, P.-C., & Hechl, S. (2021). Digital interfaces of historical newspapers: opportunities, restrictions and recommendations. The Journal of Data Mining & Digital Humanities (in press). https://zenodo.org/record/4446818#.YCPNNOgzY2w

Piotrowski, M. (2012). Natural Language Processing for Historical Texts. Morgan & Claypool Publishers.

Salmi, H., Paju, P., Rantala, H., Nivala, A., Vesanto, A., & Ginter, F. (2020). The reuse of texts in Finnish newspapers and journals, 1771–1920: A digital humanities perspective. Historical Methods: A Journal of Quantitative and Interdisciplinary History, DOI: 10.1080/01615440.2020.1803166

Savoy, J., & Naji, N. (2011). Comparative Information Retrieval Evaluation for Scanned Documents. In Proceedings of the 15th WSEAS International Conference on Computers (pp. 527–534).

Strange, C., McNamara, D., Wodak, J., & Wood, I. (2014). Mining for the Meanings of a Murder: The Impact of OCR Quality on the Use of Digitized Historical Newspapers. Digital Humanitites Quarterly, 8.

van Strien, D., Beelen, K., Ardanuy, M.C., Hosseini, K. McGillivray, B., & Colavizza, G. (2020). Assessing the Impact of OCR Quality on Downstream NLP Tasks. https://www.staff.universiteitleiden.nl/binaries/content/assets/governance-and-global-affairs/isga/artidigh_2020_7_cr.pdf

Suomen historian pikkujättiläinen (1989). Seppo Zetterberg et al. (eds). WSOY.

Taghva, K., Borsack, J., & Condit, A. (1996). Evaluation of Model-Based Retrieval Effectiveness with OCR Text. ACM Transactions on Information Systems, 14(1): 64–93. DOI: https://doi.org/10.1145/214174.214180

Tanner, S., Munoz, T. & Ros, P.H. (2009). Measuring Mass Text Digitization Quality and Usefulness. Lessons Learned from Assessing the OCR Accuracy of the British Library's 19th Century Online Newspaper Archive. D-Lib Magazine, 15(7/8). ISSN 1082-9873. DOI: https://doi.org/10.1045/july2009-munoz

Traub M.C., van Ossenbruggen J., & Hardman L. (2015). Impact Analysis of OCR Quality on Research Tasks in Digital Archives. In: Kapidakis S., Mazurek C., Werla M. (eds) Research and Advanced Technology for Digital Libraries. TPDL 2015. Lecture Notes in Computer Science, vol 9316. Springer, Cham. https://doi.org/10.1007/978-3-319-24592-8_19

Traub, M.C, van Ossenbruggen, J.R, Samar, T, & Hardman, L. (2018). Impact of Crowdsourcing OCR Improvements on Retrievability Bias. In ACM International Conference Proceeding Series. doi:10.1145/3197026.3197046

# Appendix 1. Pre-formulated queries

List of the pre-formulated queries

| ID | Query in Finnish | Rough translation |
|---|---|---|
| 1 | Bobrikoffin murha 1904 | Murder of (Nikolai) Bobrikoff in 1904 |
| 2 | Postimanifesti 1890 | Postal manifest in 1890 |
| 3 | Nuorsuomalaisen puolueen perustaminen vuonna 1894 | Founding of the young Finns' party in 1894 |
| 4 | Helmikuun manifesti 1899 | The February manifest in 1899 |
| 5 | Eduskuntavaalit 1907 | Parliamentary elections in 1907 |
| 6 | Hannes Kolehmainen Tukholman olympialaisissa 1912 | Hannes Kolehmainen at the Stockholm Olympics in 1912 |
| 7 | Maailmansodan rauha 1918 | Peace of the WWI in 1918 |
| 8 | Nansenin matka pohjoisnavalle | Nansen's expedition to the North Pole |
| 9 | Lokakuun vallankumous Venäjällä 1917 | October revolution in Russia year 1917 |
| 10 | Saksan keisarikunta 1871 | The German Empire 1871 |
| 11 | Norjan itsenäisyys 1905 | Independence of Norway in 1905 |
| 12 | Tampereen valloitus 1918 | Conquest of Tampere in 1918 |
| 13 | Suomen kuningas Friedrich Karl | Karl Friedrich, the King of Finland |
| 14 | Tokoin senaatti 1917 | The senate of (Oskari) Tokoi |
| 15 | Tukholman olympialaiset 1912 | The Olympic games of Stockholm in 1912 |
| 16 | Maamieskoulu | Agricultural school |
| 17 | Laukon torpparilakko | Sharecroppers' strike in Laukko |
| 18 | Helsingin valtaus 1918 | Occupation of Helsinki in 1918 |
| 19 | Suomen itsenäisyys 1917 | Independence of Finland in 1917 |
| 20 | Espanjantauti | The Spanish flu |
| 21 | Viaporin kapina 1906 | Rebellion in Viapori in 1906 |
| 22 | Laulaja Aino Ackte | Singer Aino Ackte |
| 23 | Suomen laulu kuoro | The choir of Finnish song |
| 24 | Suomen Naisyhdistys | The Finnish Womens' association |
| 25 | Lontoon olympialaiset 1908 | London Olympics 1908 |
| 26 | Raitiotie Helsingissä | Tramway in Helsinki |
| 27 | J. L. Runebergin kuolema 1877 | Death of J.L. Runeberg in 1877 |
| 28 | Mannerheim valtionhoitajana 1918 | (General) Mannerheim as a regent in 1918 |
| 29 | Torpparilaki | The sharecropper law |
| 30 | Elinkeinovapaus 1879 | Freedom of livelihood in 1879 |